\pdfoutput=1
\documentclass[runningheads]{llncs}

\usepackage{eccv}

\usepackage{eccvabbrv}
\usepackage{graphicx}
\usepackage{booktabs}
\usepackage{wrapfig}
\usepackage{placeins}

\usepackage[accsupp]{axessibility}
\usepackage{hyperref}
\usepackage{orcidlink}
\usepackage{colortbl}

\newcommand{\num}[1]{#1}
\newcommand{\sysname}{\emph{Chevron}}
\newcommand{\methodname}{\emph{CheXtra}}
\newcommand{\cipm}[2]{\ensuremath{{#1}_{\text{\tiny\color{gray}\,$\pm$#2}}}}
\newcommand{\cipmb}[2]{\ensuremath{{\mathbf{#1}}_{\text{\tiny\color{gray}\,$\pm$#2}}}}
\graphicspath{{figures/}}

\begin{document}

\author{Constantin Seibold, David Vinu, Matthias A. Fink}
\authorrunning{Seibold et al.}
\institute{Diagnostic and Interventional Radiology, University Clinic Heidelberg,  Germany}

\title{Building Foreign Object Segmentation datasets for Chest X-Rays with Synthetic-trained proposals} \titlerunning{Building Foreign Object Segmentation datasets for Chest X-Rays}

\maketitle

\begin{abstract}
Foreign objects and support devices such as catheters, lines, retained surgical items, and
ingested objects are among the most common abnormal findings on chest radiographs, and their
misplacement is frequent and time-critical. Support lines are malpositioned in 3--14\% of cases.
Pixel-level masks for these objects barely exist, and every cheap label-free route (i.e. foundation
segmenters, unsupervised discovery, ...) breaks on thin, rare,
or unmentioned devices. We train \methodname{},
a class-agnostic object-proposal generator that locates foreign-object regions for a human to
verify and refine. We composite procedurally rendered foreign objects onto real, device-free
radiographs, which yields exact masks for diverse appearances. Against thirteen off-the-shelf, unsupervised,
and cross-domain proposers on two unseen real chest-X-ray datasets, \methodname{} is the only
label-free method working on both domains, at Average Recall $0.50$ and $0.58$. The strongest
baseline reaches $0.72$ on one domain and falls to $0.05$ on the other. At the few-proposal
budget a curator uses, \methodname{} recalls thin catheters at $0.44$, against $0.01$ or below
for every off-the-shelf method. Ablations identify device-appearance diversity as the dominant transfer factor, with no
consistent benefit from anatomical placement realism and limited impact of corpus scaling. A human-in-the-loop engine,
\sysname{}, allows reassembling of fragmentations into masks and class-assignment of proposals by clustering, 
retrieval and manual curation. We release the code for synthetic data generation at \href{https://github.com/ConstantinSeibold/CXR\_SynthFB}{https://github.com/ConstantinSeibold/CXR\_SynthFB} and \sysname{} at \href{https://github.com/ConstantinSeibold/chevron-data-curator}{https://github.com/ConstantinSeibold/chevron-data-curator}.  
\end{abstract}

\section{Introduction}
\label{sec:intro}

\begin{figure}[t]
  \centering
  \captionsetup{belowskip=-8pt}
  \includegraphics[width=\linewidth]{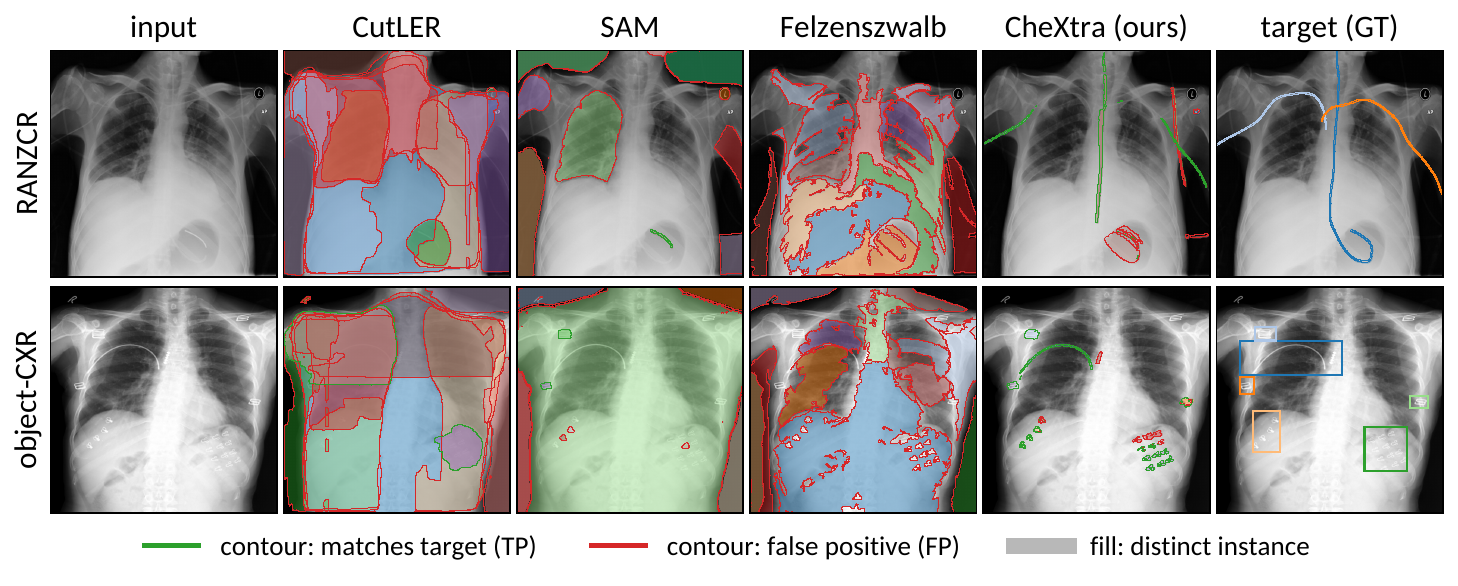}
  \caption{Class-agnostic proposal masks on unseen real chest X-rays, the top \num{12} by score:
  input, off-the-shelf proposers, ours, and target. Fill colour separates instances, and contour
  marks correctness, green for true positives and red for false positives. Baselines flood or miss
  the objects. Ours recalls thin catheters on RANZCR (top) and external objects on object-CXR
  (bottom). object-CXR targets are boxes, and all predictions are masks.}
  \label{fig:teaser}
\end{figure}

Foreign objects are among the most common abnormal findings on chest radiographs, spanning
iatrogenic support devices and lines, retained surgical items, and ingested
objects~\cite{hunter2004devices,manzella2009gossypiboma,pugmire2015ingested,roy2023analyzing}.
Getting them right is time-critical. Support lines are malpositioned in 3--14\% of cases and
central venous catheters in up to 30\%~\cite{gupta2014postprocedural, sakthivel2020malpositioned}, and a swallowed button
battery can injure the esophagus within hours~\cite{pugmire2015ingested}. Yet pixel-level
segmentation labels barely exist, and the few public resources each cover a narrow slice,
catheters and lines~\cite{ranzcr} or external objects~\cite{objectcxr}. Our target is the upstream
bottleneck, building these labels efficiently.

None of the cheap, label-free routes closes this gap.
Promptless foundation segmenters~\cite{sam,medsam} and unsupervised object
discovery~\cite{cutler} target compact, salient objects, so they miss thin, low-contrast devices
(\cref{fig:teaser}). Labels mined from radiology reports omit devices that go unmentioned. Text-conditioned generative
models~\cite{roentgen,chexgen} are difficult to adapt to this task.
A device outside the model's training distribution and text vocabulary, such as an endobronchial
coil, might not be drawn reliably and leaves no mask to extract. A detector trained on one real device dataset also does not
transfer to another (\cref{sec:proposals}).

We propose the use of synthetic data to train a class-agnostic object-proposal
generator whose task is to recall foreign-object regions of any class. We composite procedurally
rendered devices onto real, foreign-body-free chest radiographs. Only the inserted pixels are
synthetic, so every target carries an exact mask at no labeling cost. We aim for images that teach
a network to recognize non-anatomical structures in real data, so the proposer recalls shapes
off-the-shelf methods miss, rather than aiming directly for synthetic realism.

A class-agnostic proposal is not yet a label. We pair the proposer with a human-in-the-loop
curation engine that acts on groups of proposals in an embedding space instead of on single
pixels. It embeds and clusters instances, so the curator works on whole clusters at once. The curator can semi-automatically
merge the fragmented pieces of one device within a cluster. It assigns classes by nearest-neighbor
retrieval against a small labeled reference bank~\cite{segsort,protoseg}. The curator verifies and corrects in-bulk instead of drawing masks from scratch.

In this regard, we investigate which property of synthetic data
transfers to real foreign-object recall.  While placement realism and corpus scale have seemingly limited influence on the retrieval performance,   increasing the appearance diversity of synthetic generation shows noticeable improvements. (\cref{sec:proposals}). As such, we see our
contributions as follows:
\begin{itemize}
\item A compositing pipeline that inserts diverse synthetic foreign-object appearances onto real
chest radiographs with exact instance masks, and training a class-agnostic proposal generator on
them (\cref{sec:generation}).
\item A generalization study of these proposals against 13 off-the-shelf, unsupervised, and
cross-domain baselines paired with analysis on related design choices (\cref{sec:proposals}).
\item A human-in-the-loop engine that reassembles fragmented proposals into masks and assigns
classes by cluster-then-merge (\cref{sec:curation}). 
\end{itemize}

\section{Related Work}
\label{sec:related}

\noindent\textbf{Synthetic data for training-set construction.}
Compositing foreground objects onto backgrounds is a standard way to synthesize labeled data, by
cut-and-paste~\cite{dwibedi}, probability-map-guided pasting~\cite{instaboost}, or context-free
copy-paste~\cite{ghiasi}, and in medical imaging by relocating a tumor with its mask~\cite{tumorcp}
or mixing labeled and unlabeled crops~\cite{bcp}. These paste a real crop. A complementary trend
synthesizes the foreground, sourcing instances from a diffusion model~\cite{xpaste} or generating
whole chest X-rays~\cite{roentgen,cheff,chexgen}. Our generator sits between the two. It pastes
procedurally rendered foreign objects as well as real crops onto real,
device-free radiographs, which gives exact masks for diverse appearances, including rare ones
generative models do not reliably produce. We train a recall-oriented proposal generator, and we
revisit the synthetic-to-real gap~\cite{tobin,tremblay} that motivates this choice in
\cref{sec:proposals}.

\noindent\textbf{Foreign objects in chest X-rays.}
Clinically, foreign objects are diverse~\cite{hunter2004devices,manzella2009gossypiboma,pugmire2015ingested,roy2023analyzing},
but computational resources are narrow and label-coarse. object-CXR gives box, ellipse, and polygon
localization of external objects~\cite{objectcxr}, RANZCR CLiP gives catheter and line position
classification~\cite{ranzcr}, PadChest gives image-level device labels~\cite{padchest}, and
purpose-built detectors give a few boxed categories that are not publicly
released~\cite{kufel2023foreign}. None offers pixel-level instance masks. We take a step toward
that with our proposal pipeline and curation engine.

\noindent\textbf{Class-agnostic proposals and curation.}
Object proposals range from classical selective search~\cite{selectivesearch}, edge
boxes~\cite{edgeboxes}, graph-based~\cite{felzenszwalb} and combinatorial grouping~\cite{mcg} to
learned candidates~\cite{deepmask}, judged by Average Recall~\cite{hosang}. Recent work discovers
objects from self-supervised features~\cite{lost,tokencut,freesolo,cutler} or emits dense masks
from promptless foundation segmenters~\cite{sam,samhq,medsam}. A classification-free objectness
signal generalizes to unseen categories~\cite{oln}, which motivates training a class-agnostic
proposer instead of a closed-vocabulary detector, and we benchmark against these methods
(\cref{sec:proposals}). To label proposals, our curator clusters instance embeddings and retrieves
nearest neighbors, following work that segments by clustering and
retrieval~\cite{segsort,spml,protoseg,kmaxdeeplab}. Parameter-free FINCH clustering~\cite{finch}
over multi-view embeddings lets the human decide over whole clusters instead of the pixels of
click-level interactive annotation~\cite{dextr}. 

\section{Synthetic Foreign-Object Generation}
\label{sec:generation}

\begin{figure}[tb]
  \centering
  \includegraphics[width=\linewidth]{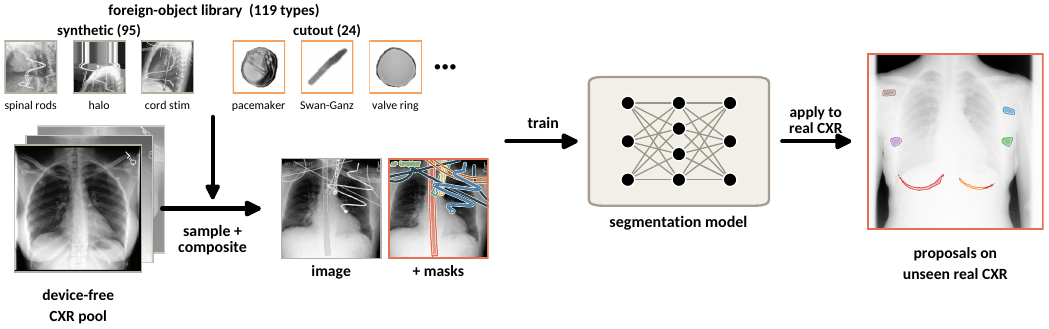}
  \caption{\textbf{Synthetic foreign-object generation.} From a library of \num{119}
  foreign-object types, procedurally rendered or cut-paste from real crops, we composite objects
  onto device-free MIMIC-CXR backgrounds with X-ray-aware rendering. Only inserted pixels are
  synthetic, so each instance mask is exact at no labeling cost. The paired masked and unmasked
  images train a class-agnostic model that then proposes foreign-object masks on unseen real chest
  X-rays (\cref{sec:proposals}).}
  \label{fig:pipeline}
\end{figure}

\noindent\textbf{Overview.}
A real, foreign-body-free chest radiograph provides the background, and only the inserted
foreign-object pixels are synthetic (\cref{fig:pipeline}). Because we draw each object, its
instance mask is exact at no labeling cost. The synthetic data only need to teach a network to recall
non-anatomical structures in real images, so we favour appearance diversity over photorealism.

\noindent\textbf{Base images.}
Backgrounds are device-free MIMIC-CXR~\cite{mimic} images. Report-derived labels are an unreliable
proxy for image content, so we select the pool by pairing report-text filtering with a learned
image-side classifier, a logistic regression on RAD-DINO~\cite{raddino} embeddings, that flags
images with residual, report-unmentioned devices.

\begin{figure}[tb]
  \centering
  \includegraphics[width=\linewidth]{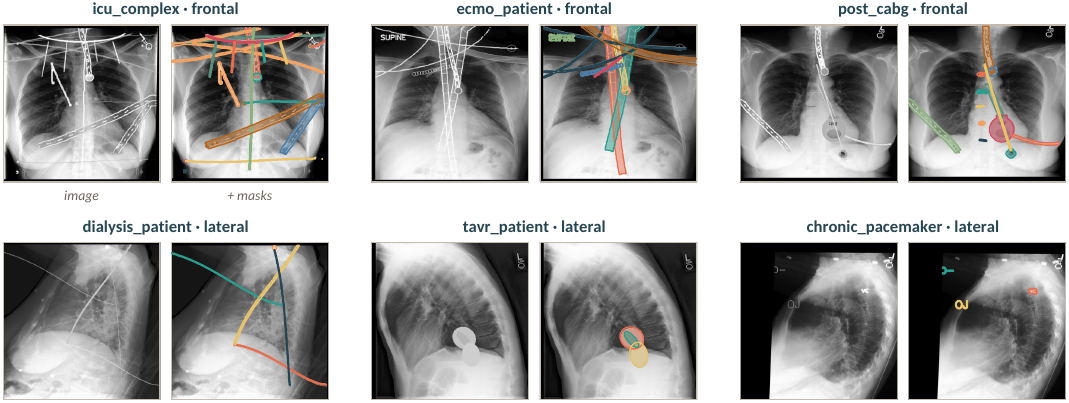}
  \caption{Example synthetic images across clinical archetypes and views, each shown without and
  with its instance-mask overlay, one colour per instance. }
  \label{fig:genexamples}
\end{figure}

\noindent\textbf{Foreign-object synthesis.}
The object library spans \num{119} categories: tubes and lines, leads and wires, orthopedic
hardware, implant bodies, stents and grafts, external surface objects, ingested items, and imaging
artifacts. Each object is drawn by a procedural shape function or cut-paste from a small set of
real device crops, then composited with an X-ray-aware rendering mode (alpha overlay, soft-edge
feather, Beer--Lambert attenuation, rib and clavicle occlusion, translucent implant) under
randomized contrast and opacity. This breadth of appearance, including rare devices that current text-conditioned generative chest-X-ray models do not reliably synthesize, teaches the proposer to recall shapes off-the-shelf methods miss
(\cref{fig:genexamples}).

\noindent\textbf{Instance sampling.} An image draws up to $k=12$ objects. With probability $p=0.85$ their categories come from one of 23 weighted clinical archetypes (e.g. ICU post-CABG, dialysis, ingestion). Each fixes a few co-occurring devices, adds its optional ones with even odds, and tops up with a small remainder drawn uniform over the whole library, so a rare implant appears as often as a common catheter. Per-category counts still respect clinical multiplicity (i.e. one pacemaker, multiple clips,\dots), and each instance is independently placed, scaled, rotated, and rendered.

\noindent\textbf{From masks to a proposer.}
We train a standard class-agnostic instance segmenter on the synthetic masks, with all categories
collapsed to one foreground class, and use it as an object-proposal generator. Its value is the
label-free, domain-transferable recall we quantify next (\cref{sec:proposals}).

\section{Experiments}
\label{sec:proposals}

\subsection{Experimental Setup}

\noindent\textbf{Proposers.}
We compare proposers spanning the study's axes. \methodname{} (ours) uses synthetic foreign bodies.
The in-domain upper bounds RANZCR-real and object-CXR-real use one real set each and test whether a
single real dataset transfers. The natural-image detectors are COCO-Mask2Former and COCO-Mask
R-CNN. CutLER is unsupervised. The promptless foundation segmenters are SAM, SAM-HQ, SAM2, and
MedSAM in automatic mask generation. Felz$\rightarrow$\allowbreak MedSAM prompts MedSAM with
untrained boxes. Felzenszwalb superpixels are classical, and Random-box is a flooding control.
Learned proposers are Mask2Former-family unless noted.

\noindent\textbf{Metric.}
We measure class-agnostic
Average Recall, $\mathrm{AR}@k$, the fraction of ground-truth instances covered by some
score-ranked top-$k$ proposal, averaged over images, which we chose as no test ground truth contains annotations for all present foreign objects. We evaluate over an IoU sweep and report 95\% bootstrap confidence
intervals over images with 1000 resamples.

\noindent\textbf{External test sets.}
\textbf{object-CXR} has a dev-set of \num{500} device-bearing images and \num{1015} class-agnostic
foreign-object boxes. Its masks are coarsely annotated, so we score box recall at IoU 0.3 instead.
For \textbf{RANZCR}, we consider a \num{1000}-image held-out sample with no public segmentation split and \num{1500} catheter and line instances. We generate the masks from the point annotations by connecting them via a 3~px line and score the mask recall at IoU 0.1. Both sets are disjoint from every in-domain model's
training data.

\begin{table}[tb]
\caption{Cross-domain generalization: class-agnostic Average Recall at the curation budget
($\mathrm{AR}@10$, $\mathrm{AR}@50$, $\mathrm{AR}@100$),
over the 500 object-CXR / 1000 RANZCR held-out images. object-CXR uses \emph{box} recall, RANZCR \emph{mask} recall. Top rows are in-domain upper
bounds; \textbf{bold} is best excluding them. Subscripts are 95\% bootstrap CI half-widths over
images (1000 resamples).}
\label{tab:headline}
\centering
\scriptsize
\setlength{\tabcolsep}{3pt}
\renewcommand{\arraystretch}{1.25}
\resizebox{\linewidth}{!}{%
\begin{tabular}{@{}l ccc >{\hspace{1.8em}}c cc@{}}
\toprule
& \multicolumn{3}{c}{object-CXR (box, IoU 0.3)} & \multicolumn{3}{c}{RANZCR (mask, IoU 0.1)}\\
\cmidrule(lr){2-4}\cmidrule(lr){5-7}
Method & $\mathrm{AR}@10$ & $\mathrm{AR}@50$ & $\mathrm{AR}@100$ & $\mathrm{AR}@10$ & $\mathrm{AR}@50$ & $\mathrm{AR}@100$\\
\midrule
\rowcolor[gray]{0.9} object-CXR-real~\cite{objectcxr} \emph{(in-dom.)}    & \cipm{0.896}{.023} & \cipm{0.943}{.017} & \cipm{0.967}{.013} & \cipm{0.024}{.008} & \cipm{0.045}{.01} & \cipm{0.059}{.011}\\
\rowcolor[gray]{0.9} RANZCR-real~\cite{ranzcr} \emph{(in-dom.)}           & \cipm{0.089}{.022} & \cipm{0.128}{.026} & \cipm{0.134}{.026} & \cipm{0.997}{.003} & \cipm{0.998}{.003} & \cipm{0.998}{.003}\\
\cmidrule(lr){1-7}
COCO-Mask2Former~\cite{mask2former}                  & \cipm{0.078}{.021} & \cipm{0.142}{.028} & \cipm{0.177}{.031} & \cipm{0.000}{.001} & \cipm{0.003}{.002} & \cipm{0.004}{.003}\\
COCO-Mask R-CNN~\cite{maskrcnn}                      & \cipm{0.123}{.026} & \cipm{0.274}{.035} & \cipm{0.356}{.037} & \cipm{0.003}{.003} & \cipm{0.009}{.005} & \cipm{0.017}{.006}\\
SAM~\cite{sam}                                       & \cipm{0.136}{.025} & \cipm{0.194}{.028} & \cipm{0.194}{.028} & \cipm{0.011}{.005} & \cipm{0.022}{.007} & \cipm{0.022}{.007}\\
SAM-HQ~\cite{samhq}                                  & \cipm{0.149}{.027} & \cipm{0.209}{.03} & \cipm{0.209}{.03} & \cipm{0.013}{.005} & \cipm{0.034}{.009} & \cipm{0.034}{.009}\\
SAM2~\cite{sam2}                                     & \cipm{0.118}{.025} & \cipm{0.120}{.025} & \cipm{0.120}{.025} & \cipm{0.000}{.000} & \cipm{0.000}{.000} & \cipm{0.000}{.000}\\
MedSAM~\cite{medsam}                                 & \cipm{0.110}{.025} & \cipm{0.269}{.034} & \cipm{0.387}{.035} & \cipm{0.002}{.002} & \cipm{0.015}{.006} & \cipm{0.026}{.008}\\
CutLER~\cite{cutler}                                 & \cipm{0.188}{.031} & \cipmb{0.579}{.037} & \cipmb{0.716}{.034} & \cipm{0.001}{.001} & \cipm{0.026}{.007} & \cipm{0.046}{.009}\\
\cmidrule(lr){1-7}
Felzenszwalb~\cite{felzenszwalb}                     & \cipm{0.133}{.026} & \cipm{0.318}{.035} & \cipm{0.408}{.035} & 0.000 & \cipm{0.236}{.023} & \cipm{0.376}{.027}\\
Selective Search~\cite{selectivesearch}              & \cipm{0.084}{.023} & \cipm{0.252}{.035} & \cipm{0.453}{.036} & 0.000 & 0.000 & 0.000\\
Felz$\rightarrow$MedSAM~\cite{felzenszwalb,medsam}   & \cipm{0.088}{.022} & \cipm{0.238}{.032} & \cipm{0.334}{.036} & \cipm{0.005}{.003} & \cipm{0.043}{.01} & \cipm{0.083}{.013}\\
Random-box                                           & \cipm{0.111}{.022} & \cipm{0.352}{.035} & \cipm{0.504}{.037} & 0.000 & 0.000 & 0.000\\
\cmidrule(lr){1-7}
\methodname{} (ours)                                       & \cipmb{0.377}{.038} & \cipm{0.486}{.039} & \cipm{0.503}{.039} & \cipmb{0.436}{.028} & \cipmb{0.548}{.026} & \cipmb{0.580}{.026}\\
\bottomrule
\end{tabular}%
}
\end{table}

\noindent\textbf{Implementation and protocol.}
Every proposer we train, that is \methodname{}, the two in-domain models, and the ablation variants
below use a Mask2Former~\cite{mask2former} decoder with \num{200} queries on a
COCO~\cite{coco}-initialized DINOv3 ViT-S/16~\cite{dinov3} backbone, fine-tuned at $512$\,px for
\num{6} epochs with batch \num{4} and AdamW.  They train on native label spaces and
are scored class-agnostically. Off-the-shelf baselines run at public defaults. \methodname{} and other trained versions run at
about 0.2\,s per image, an order of magnitude faster than SAM-family generators at 1 to 1.7\,s.

Each image is resized to its ground-truth $512$\,px frame, so proposals and ground truth share
coordinates. We report score-ranked $\mathrm{AR}@k$ for $k\in\{10, 50,100\}$ at the  box IoU $0.3$ for object-CXR and mask IoU $0.1$ for RANZCR thresholds and show a full $k$ sweep in \cref{fig:curves}. A per-IoU sweep over box
$0.3$ to $0.7$ and mask $0.1$ to $0.5$ is in the \cref{tab:periou}. All outputs are
scored class-agnostically with one foreground class.

\noindent\textbf{Data provenance and licensing.}
Backgrounds are MIMIC-CXR~\cite{mimic} under the PhysioNet Credentialed Health Data License, used
under credentials. We redistribute no MIMIC pixels or free-text reports. The
object-CXR~\cite{objectcxr} and RANZCR~\cite{ranzcr} public sets serve only as evaluation
references, disjoint from every in-domain model's training data (\cref{sec:proposals}). The curated
masks are a single-annotator demonstration of the engine and not a released benchmark. It is not intended for clinical use.

\subsection{Results}
\noindent\textbf{Cross-domain generalization.}
We show the quantitative performance of the individual mask-proposal methods in \Cref{tab:headline}. We see that current real datasets do not transfer. RANZCR-real
reaches 0.998 in domain but 0.134 on object-CXR, and object-CXR-real reaches 0.967 in domain but
0.059 on RANZCR. Among label-free methods, \methodname{} is the strongest on both domains, at 0.503
and 0.580. The methods that rival it on one domain collapse on the other. CutLER reaches 0.716 on
object-CXR and 0.046 on RANZCR. The only other proposer that avoids collapse, Felzenszwalb, trails
\methodname{} on each domain, at 0.408 and 0.376. 

\begin{figure}[tb]
  \centering
  \includegraphics[width=\linewidth]{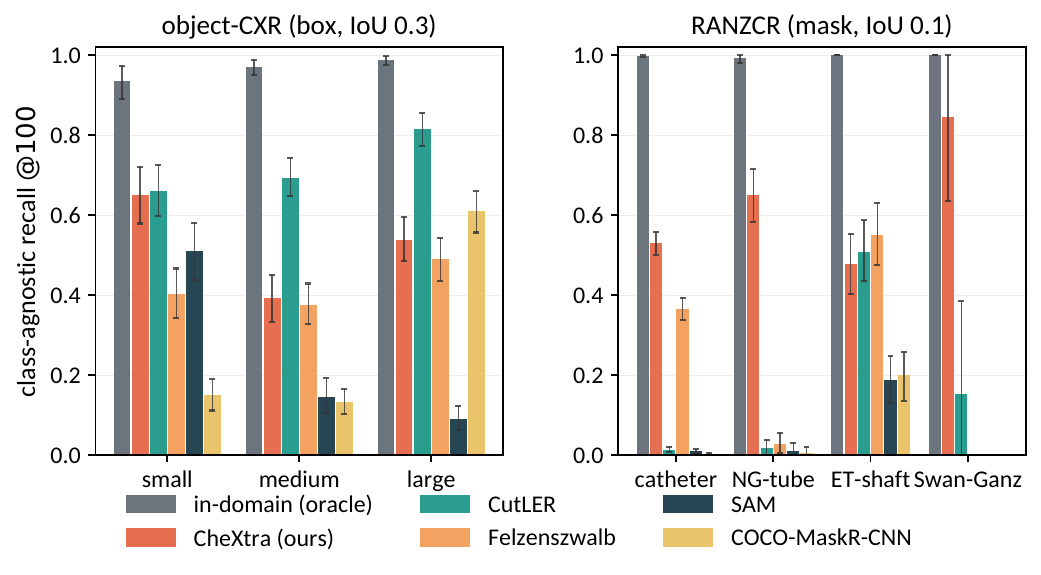}
  \caption{\textbf{Breakdown by object size and line type.} Class-agnostic recall, $\mathrm{AR}@100$
  at the operating IoU, split by object size on object-CXR and line type on RANZCR, with 95\% CI
  error bars. }
  \label{fig:persize}
\end{figure}

\noindent\textbf{Breakdown by object size and line type.}
By object size (\cref{fig:persize}, object-CXR box $\mathrm{AR}@100$ at IoU 0.3), \methodname{}
matches CutLER on small objects, at 0.65 against 0.66, and trails on medium and large, at 0.39 and
0.54 against 0.69 and 0.82. Per line type on RANZCR masks, \methodname{} leads every thin type, with
catheter 0.53, NG-tube 0.65, and Swan-Ganz 0.85 against 0.37 or below for the next method, and
trails only on the ET-tube shaft, at 0.48 against 0.62 for Felz$\rightarrow$\allowbreak MedSAM.
Oracle best mask IoU and Dice on RANZCR rank \methodname{} 0.148 and 0.239 above Felzenszwalb 0.098
and 0.173 above CutLER 0.050 and 0.092, against a ceiling of 0.506 and 0.660. 
Qualitatively (\cref{fig:teaser}), off-the-shelf proposers flood or miss foreign objects, and
\methodname{} recalls them on both datasets.

\begin{figure}[t]
  \centering
  \includegraphics[width=\linewidth]{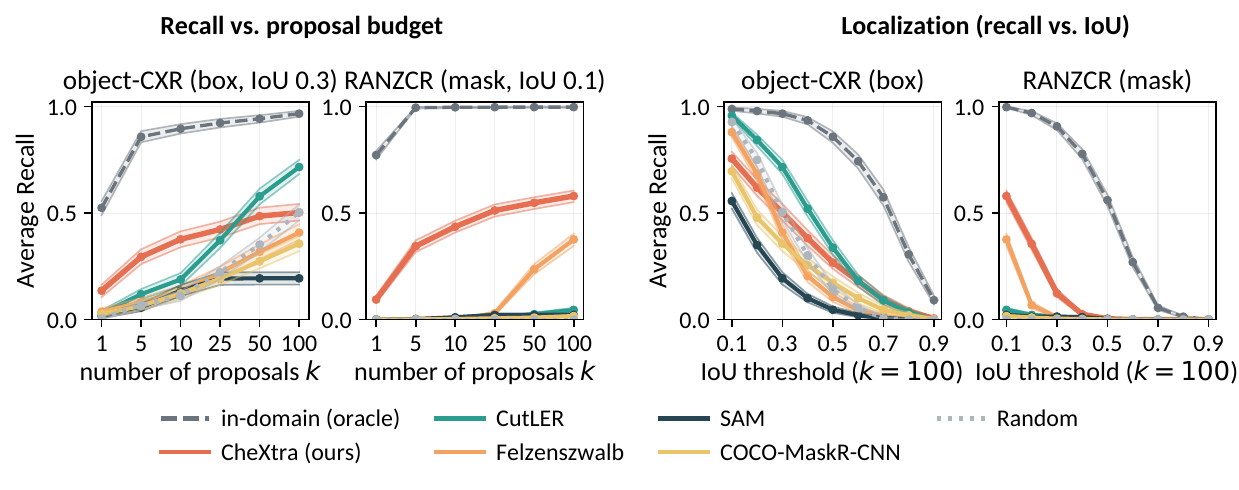}
  \caption{Class-agnostic Average Recall across domains. Bands are 95\% bootstrap CIs, and
  in-domain models bound each panel. The left pair shows recall against budget $k$ at fixed IoU. The
  right pair shows recall against IoU at $k{=}100$. }
  \label{fig:curves}
\end{figure}

\begin{figure}[t]
  \centering
  \includegraphics[width=\linewidth]{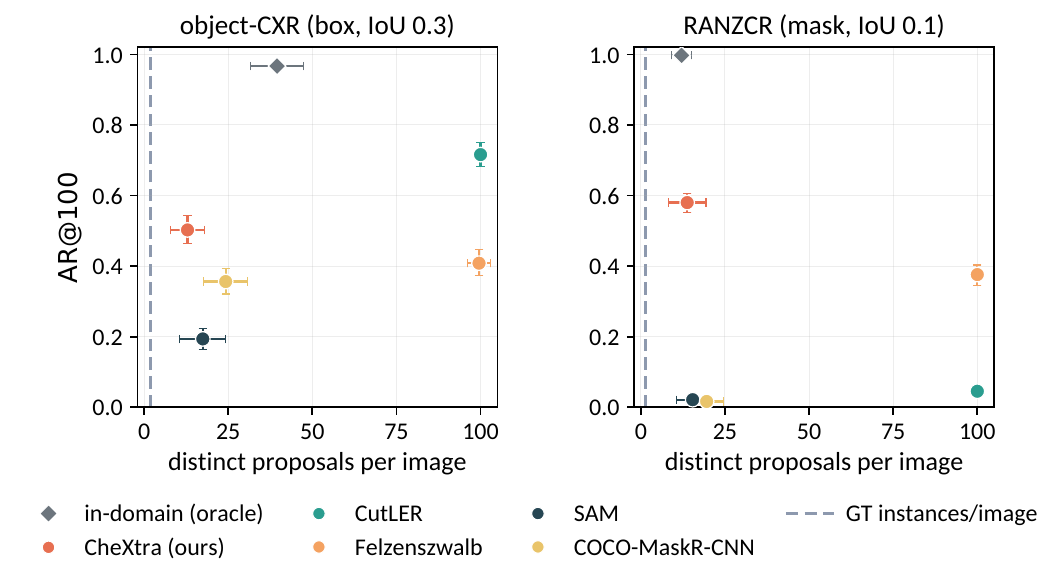}
  \caption{\textbf{Distinct proposals versus recall.} $\mathrm{AR}@100$ at the operating IoU, with
  vertical bars for 95\% CI, against distinct proposals per image after class-agnostic NMS at box
  IoU 0.5, capped at 100, with horizontal bars for $\pm 1$ SD across images. The dashed line marks
  the mean ground-truth instances per image.
  }
  \label{fig:flooding}
\end{figure}

\FloatBarrier
\noindent\textbf{Budget analysis.}
The proposal budget at which recall is measured reorders the methods (\cref{tab:headline}).
\methodname{} leads on both domains at the curation budget, reaching 0.377 against 0.188 for
CutLER on object-CXR and 0.436 against 0.013 or below for every off-the-shelf method on RANZCR
masks. It reaches that recall with about 13 proposals per image, close to the two instances an
image holds on average, while CutLER and Felzenszwalb spend the full 100 (\cref{fig:flooding}).
CutLER passes \methodname{} on object-CXR only at that full budget ($\mathrm{AR}@50$ 0.579,
$\mathrm{AR}@100$ 0.716), because its top-ranked proposals are rarely the foreign object, so its
lead is volume rather than localization. That lead erodes as the IoU tightens, reverses at IoU
0.7
(\cref{tab:periou}). A paired per-image bootstrap confirms the pattern:
\methodname{} gains \cipm{+0.189}{.039} and \cipm{+0.436}{.028} $\mathrm{AR}@10$ on object-CXR and
RANZCR, trails CutLER by \cipm{-0.213}{.047} on object-CXR at full budget, and still leads RANZCR
masks by \cipm{+0.204}{.033}. It leads thin catheter masks at every operating point
(\cref{fig:curves}).

\begin{wrapfigure}{r}{0.5\linewidth}
\vspace{-1cm}
  \centering
  \includegraphics[width=\linewidth]{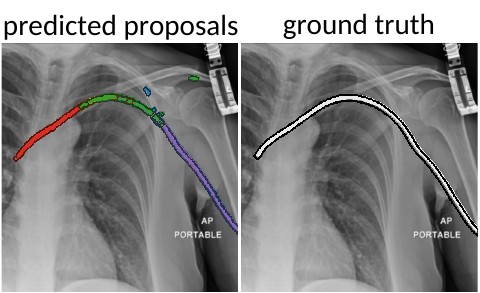}
  \caption{\textbf{Fragmentation of one device.} The proposer recalls this catheter as disjoint
  fragments (\emph{left}, distinct colours, dilated for visibility) whose union reconstructs $92\%$
  of this ground-truth instance (\emph{right}). }
  \label{fig:fragex}
  \vspace{-1cm}
\end{wrapfigure}

\noindent\textbf{Fragmentation.}
The proposer often recalls a long device as many fragments rather than one mask (\cref{fig:fragex}), so
RANZCR mask recall keeps climbing with the budget (\cref{fig:curves}), which we quantify in  \Cref{tab:fragmerge}. \methodname{} splits a RANZCR line into 4.3 proposals on average, and 9.0 for the
Swan-Ganz, and their union recovers more of the line than any single proposal, adding 0.15 pixel
recall on the Swan-Ganz. Off-the-shelf, unsupervised, and classical proposers split a line 0.7
times or fewer, so they barely fire along it and leave nothing to merge, with a merge gain of 0.02
or below. Reassembling fragments is therefore the curator's primary operation (\cref{sec:curation}).

\begin{table}[tb]
\caption{\textbf{Fragmentation and merge recovery} (RANZCR mask, top-$100$ proposals). Each GT
line's proposals are greedily assigned to the line they most overlap; we report fragments per GT
instance and the recall of GT-line pixels by the single best proposal \vs the union of its
fragments ($\Delta$ = merge gain). Only line-aware proposers fragment a device, so only they have
anything to merge. Top row is the in-domain upper bound; \textbf{bold} is best excluding it.
Subscripts are 95\% bootstrap CI half-widths over images.}
\label{tab:fragmerge}
\centering
\scriptsize
\setlength{\tabcolsep}{6pt}
\renewcommand{\arraystretch}{1.25}
\begin{tabular}{@{}l c cc c@{}}
\toprule
 & & \multicolumn{2}{c}{cov.\ of GT line} & \\
\cmidrule(lr){3-4}
 & predictions/GT & single & merged & $\Delta$\\
\midrule
RANZCR-real~\cite{ranzcr} \emph{(in-dom.)} & \cipm{16.65}{.58} & \cipm{0.726}{.011} & \cipm{0.775}{.012} & \cipm{+0.049}{.002}\\
\cmidrule(lr){1-5}
COCO-Mask2Former~\cite{mask2former}        & \cipm{0.01}{.01} & \cipm{0.001}{.002} & \cipm{0.001}{.002} & +0.000\\
SAM~\cite{sam}                             & \cipm{0.04}{.01} & \cipm{0.021}{.006} & \cipm{0.021}{.006} & +0.000\\
CutLER~\cite{cutler}                       & \cipm{0.31}{.04} & \cipm{0.024}{.005} & \cipm{0.025}{.006} & +0.001\\
Felzenszwalb~\cite{felzenszwalb}           & \cipm{0.74}{.05} & \cipm{0.070}{.006} & \cipm{0.087}{.007} & \cipm{+0.017}{.002}\\
\cmidrule(lr){1-5}
\methodname{} (ours)                             & \cipmb{4.31}{.44} & \cipmb{0.162}{.01} & \cipmb{0.198}{.013} & \cipmb{+0.036}{.005}\\
\bottomrule
\end{tabular}
\end{table}

\begin{table}[ht]
\caption{IoU robustness: class-agnostic $\mathrm{AR}@100$ at three IoU thresholds over the 500
object-CXR / 1000 RANZCR held-out films. Top rows are in-domain upper bounds; \textbf{bold} is
best excluding them. Subscripts are 95\% bootstrap CI half-widths over images.}
\label{tab:periou}
\centering
\scriptsize
\setlength{\tabcolsep}{3pt}
\renewcommand{\arraystretch}{1.25}
\resizebox{\linewidth}{!}{%
\begin{tabular}{@{}l ccc @{\hskip 1.8em} ccc@{}}
\toprule
& \multicolumn{3}{c}{object-CXR (box, $\mathrm{AR}@100$)} & \multicolumn{3}{c}{RANZCR (mask, $\mathrm{AR}@100$)}\\
\cmidrule(lr){2-4}\cmidrule(lr){5-7}
trained on $\rightarrow$ eval & IoU 0.3 & IoU 0.5 & IoU 0.7 & IoU 0.1 & IoU 0.3 & IoU 0.5\\
\midrule
object-CXR-real~\cite{objectcxr} \emph{(in-dom.)}    & \cipm{0.967}{.013} & \cipm{0.859}{.026} & \cipm{0.574}{.037} & \cipm{0.059}{.011} & \cipm{0.001}{.001} & 0.000\\
RANZCR-real~\cite{ranzcr} \emph{(in-dom.)}           & \cipm{0.134}{.026} & \cipm{0.021}{.01} & 0.000 & \cipm{0.998}{.003} & \cipm{0.907}{.015} & \cipm{0.561}{.027}\\
\cmidrule(lr){1-7}
COCO-Mask2Former~\cite{mask2former}                  & \cipm{0.177}{.031} & \cipm{0.089}{.023} & \cipm{0.022}{.012} & \cipm{0.004}{.003} & \cipm{0.001}{.001} & \cipm{0.001}{.001}\\
COCO-Mask R-CNN~\cite{maskrcnn}                      & \cipm{0.356}{.037} & \cipm{0.173}{.03} & \cipm{0.046}{.016} & \cipm{0.017}{.006} & \cipm{0.004}{.003} & \cipm{0.001}{.001}\\
SAM~\cite{sam}                                       & \cipm{0.194}{.028} & \cipm{0.046}{.014} & \cipm{0.007}{.006} & \cipm{0.022}{.007} & \cipm{0.014}{.005} & \cipm{0.003}{.002}\\
SAM-HQ~\cite{samhq}                                  & \cipm{0.209}{.03} & \cipm{0.064}{.019} & \cipm{0.021}{.011} & \cipm{0.034}{.009} & \cipm{0.016}{.006} & \cipm{0.004}{.002}\\
SAM2~\cite{sam2}                                     & \cipm{0.120}{.025} & \cipm{0.039}{.015} & \cipm{0.006}{.006} & \cipm{0.000}{.000} & \cipm{0.000}{.000} & \cipm{0.000}{.000}\\
MedSAM~\cite{medsam}                                 & \cipm{0.387}{.035} & \cipm{0.131}{.025} & \cipm{0.018}{.01} & \cipm{0.026}{.008} & 0.000 & 0.000\\
CutLER~\cite{cutler}                                 & \cipmb{0.716}{.034} & \cipmb{0.338}{.035} & \cipm{0.090}{.022} & \cipm{0.046}{.009} & \cipm{0.014}{.005} & \cipm{0.002}{.002}\\
\cmidrule(lr){1-7}
Felzenszwalb~\cite{felzenszwalb}                     & \cipm{0.408}{.035} & \cipm{0.106}{.023} & \cipm{0.017}{.01} & \cipm{0.376}{.027} & \cipm{0.012}{.006} & 0.000\\
Selective Search~\cite{selectivesearch}              & \cipm{0.453}{.036} & \cipm{0.173}{.028} & \cipm{0.048}{.018} & 0.000 & 0.000 & 0.000\\
Felz$\rightarrow$MedSAM~\cite{felzenszwalb,medsam}   & \cipm{0.334}{.036} & \cipm{0.071}{.019} & \cipm{0.012}{.008} & \cipm{0.083}{.013} & \cipm{0.004}{.002} & \cipm{0.000}{.001}\\
Random-box                                           & \cipm{0.504}{.037} & \cipm{0.143}{.025} & \cipm{0.007}{.005} & 0.000 & 0.000 & 0.000\\
\cmidrule(lr){1-7}
\methodname{} (ours)                                       & \cipm{0.503}{.039} & \cipm{0.268}{.033} & \cipmb{0.093}{.022} & \cipmb{0.580}{.026} & \cipmb{0.123}{.019} & \cipm{0.004}{.003}\\
\bottomrule
\end{tabular}%
}
\end{table}

\subsection{Ablations}

\begin{figure}[tb]
  \centering
  \includegraphics[width=\linewidth]{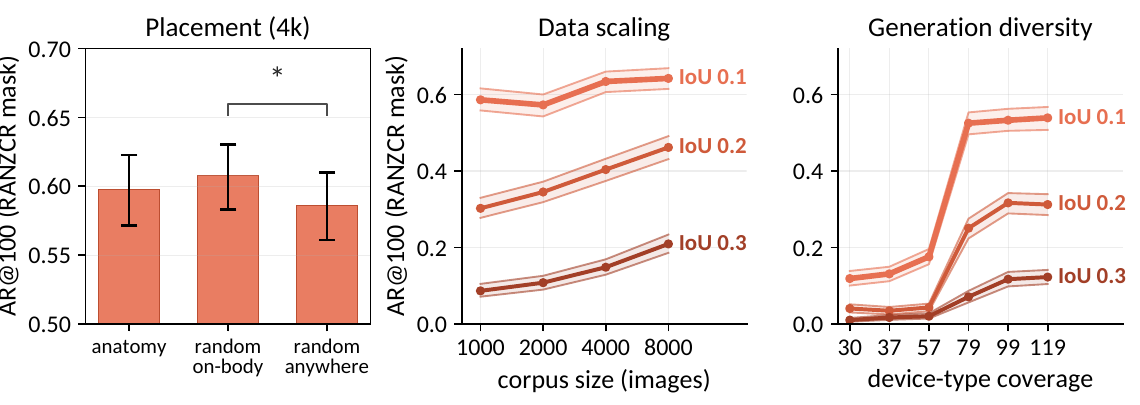}
  \caption{Ablations on RANZCR mask, mean with 95\% CI. \emph{Left:} placement $\mathrm{AR}@100$,
  with $y$ zoomed to $[0.5,0.70]$ and the arms near-equal. A paired bootstrap finds only
  random-on-body above random-anywhere significant ($*$, $p{=}0.001$). \emph{Middle and right:} $\mathrm{AR}@100$
  against corpus size and device-type coverage, with one line per match IoU and darker lines for
  stricter IoU. Gains concentrate at stricter IoU. From 1k to 8k, IoU-0.3 recall rises from 0.09 to
  0.21 while IoU-0.1 rises only from 0.59 to 0.64. }
  \label{fig:scaling}
\end{figure}

\noindent\textbf{Ablation setup.}
We reuse the Mask2Former with DINOv3 backbone proposer from above and  train on the resulting corpus in question. Placement toggles only where objects go, along a ladder of decreasing anatomical prior with 4k generated images:
anatomy from CXAS\cite{seibold2023accurate, Seibold_2022_BMVC} regions, then random on-body with a $\pm10\%$ shift, then random anywhere with a $\pm40\%$ shift and off-body placement. For data scaling, we perform corpus size sweeps 1k to 8k images at 8 epochs each. Device
diversity keeps a fraction of the \num{119} generation types.

\noindent\textbf{Placement.}
The placement panel compares three arms at 4k images in (\cref{fig:scaling}, left). A paired bootstrap separates just one pair, random on-body above random anywhere
($p{=}0.001$). Anatomy overlaps random on-body
and is not distinguishable from it, so placing objects at anatomically correct sites
adds nothing beyond keeping them on the body.

\noindent\textbf{Corpus size and diversity.}
We show the impact of the generated corpus size in (\cref{fig:scaling}). From 1k to 8k images, RANZCR mask
$\mathrm{AR}@100$ rises from 0.59 to 0.64. The gain
concentrates at stricter localization, where RANZCR IoU-0.3 recall rises from 0.09 to 0.21, so extra
data has more impact on improving masks than it influences a widening of coverage. Generation diversity has the larger effect and
acts as a threshold. Below half of the device types (30, 37, and 57 of 119), RANZCR mask recall sits at a low floor of 0.12, 0.13, and 0.18. It jumps to 0.53 at 79 of 119 types and saturates at 0.54 at full coverage.

\begin{figure}[tb]
  \centering
  \setlength{\tabcolsep}{2pt}
  \begin{tabular}{cc}
    \includegraphics[width=0.49\linewidth, height=0.35\linewidth]{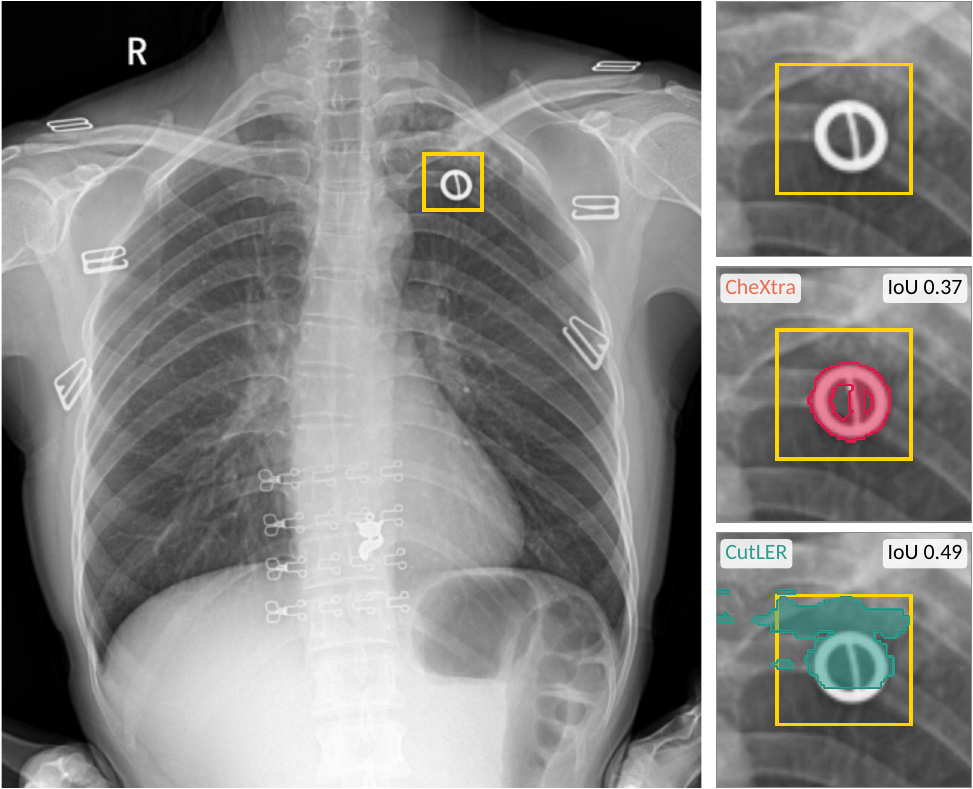} &
    \includegraphics[width=0.49\linewidth, height=0.35\linewidth]{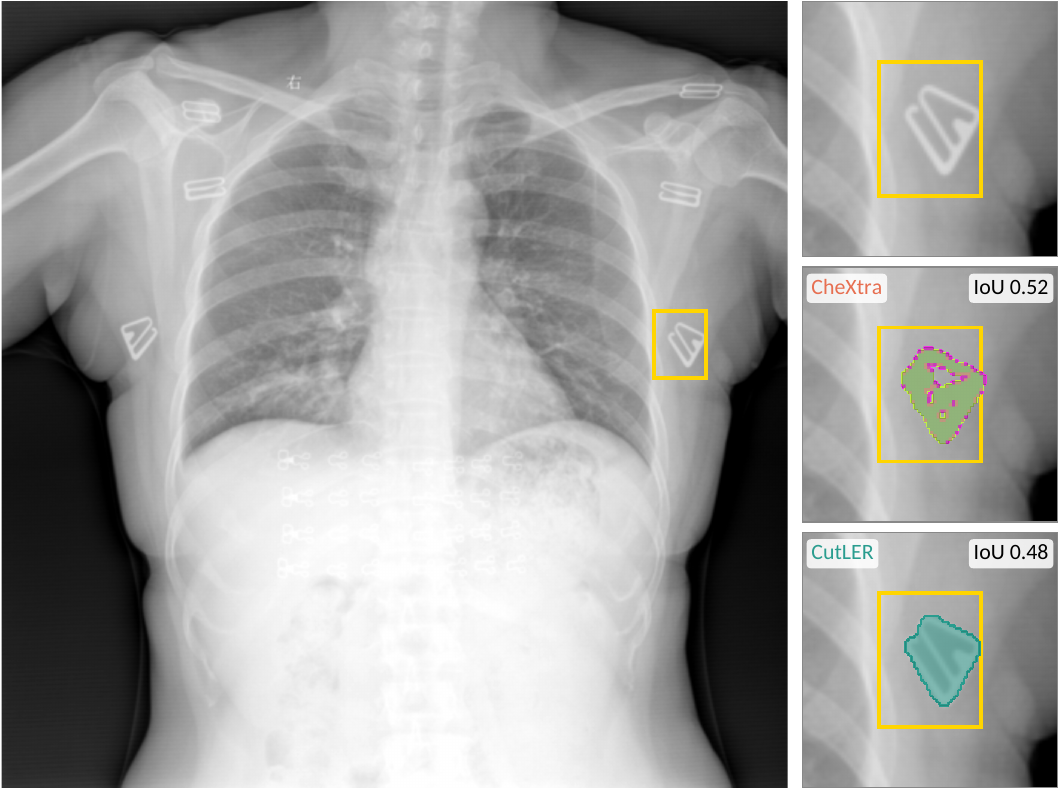} \\
  \end{tabular}
  \caption{\textbf{Annotation ceiling on object-CXR.} Annotated box (yellow) with zoomed target, CheXtra and CutLER predictions, labelled by box IoU. CheXtra follows the object outline yet scores 0.37 and 0.52 against the loose box, below CutLER's less precise 0.49 in the first case, so the recall understates mask quality.}
  \label{fig:qual-annotation-ceiling}
\end{figure}

\subsection{Qualitative analysis}
We complement the ablations with two qualitative panels that highlight (i) how close synthetic-only training gets to curated supervision, and (ii) how a generic label-free proposer fails under domain shift.

\label{sec:qual}

\noindent\textbf{Annotation ceiling on object-CXR.}
object-CXR marks each foreign object with a bounding box drawn loosely around it, so
scoring a mask against that box understates a correct segmentation. In \cref{fig:qual-annotation-ceiling}
CheXtra's mask follows the object outline, yet its enclosing box overlaps the annotation
at $0.37$ and $0.52$ IoU across the two examples; in the first, the tight prediction falls
below the $0.5$ threshold used to count a match. Box overlap can also favour the less
faithful mask: a CutLER region that fills the annotated box reaches $0.49$ IoU against
CheXtra's $0.37$ while covering the object less precisely. The box-level recall we report
on object-CXR (\cref{tab:headline}) is therefore a loose estimate on the quality a mask-level annotation would credit.

\noindent\textbf{Devices outside the benchmarks.}

\begin{figure}[tb]
  \centering
  \setlength{\tabcolsep}{2pt}
  \begin{tabular}{cc}
    \includegraphics[width=0.49\linewidth]{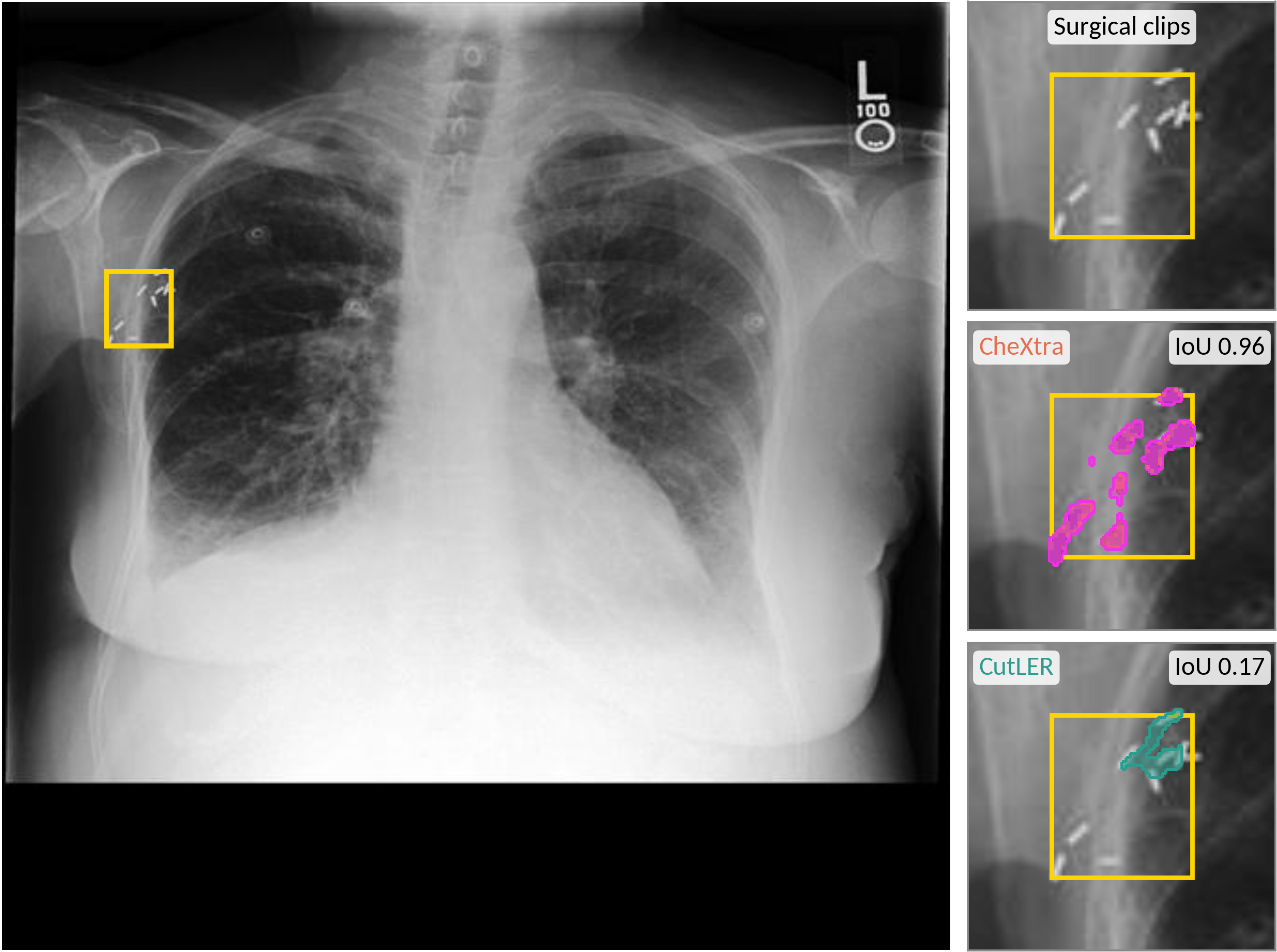} &
    \includegraphics[width=0.49\linewidth]{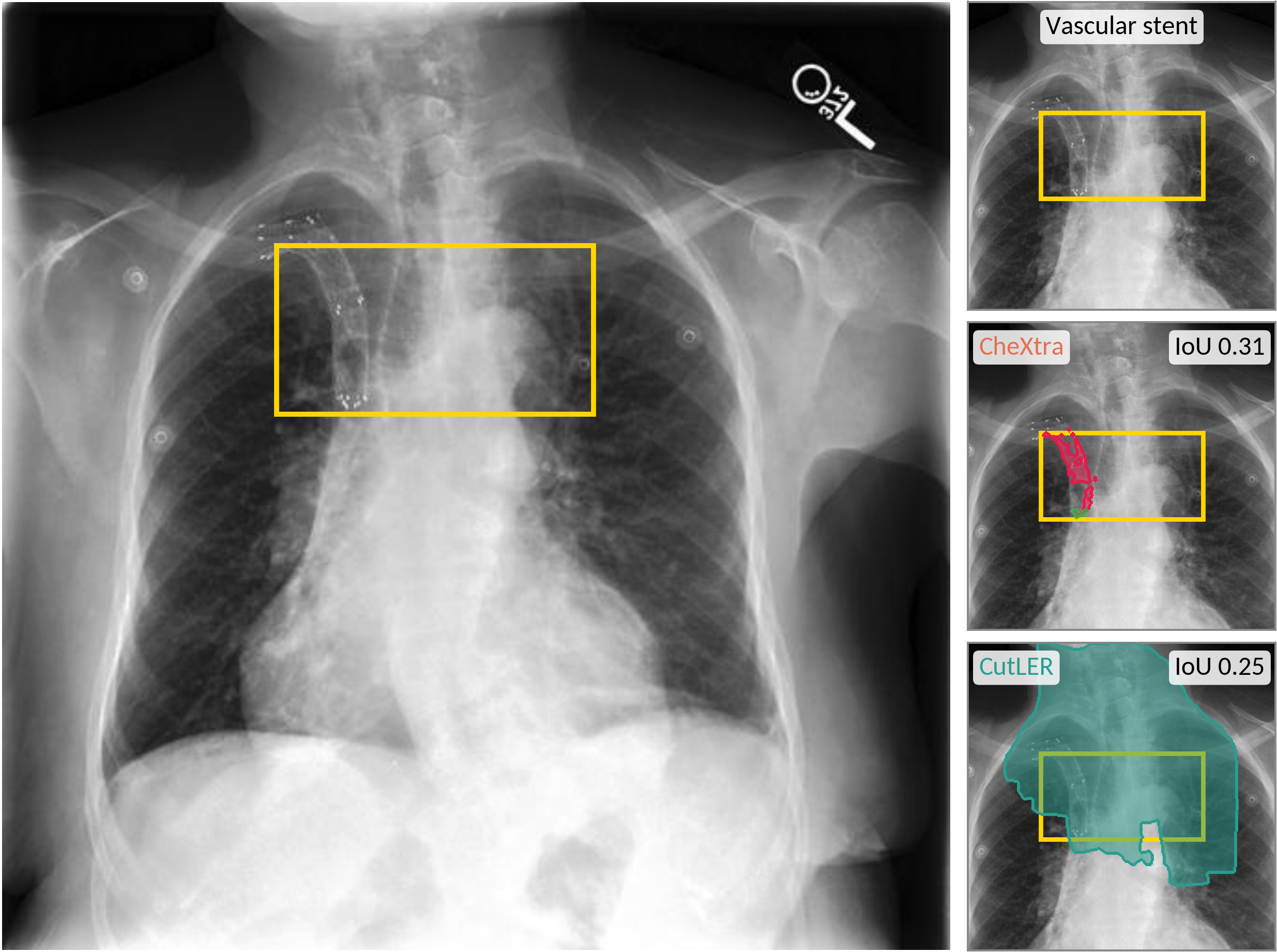} \\
  \end{tabular}
  \caption{\textbf{Qualitative OOD failures of CutLER.} A generic unsupervised objectness signal tends to favour salient blobs or anatomy-like regions and misses the thin, elongated support devices that define the long tail in chest radiographs.}
  \label{fig:qual-ood-cutler}
\end{figure}
RANZCR annotates lines and catheters and object-CXR a mixed set of objects, so neither
scores the implanted and surgical hardware common on clinical images. We indexed such cases
with the \sysname~curator (\cref{sec:curation}), selecting real MIMIC-CXR images whose foreign
objects a reviewer had labelled as non-line devices, and applied CheXtra and CutLER without
further training. \cref{fig:qual-ood-cutler} shows surgical clips and a vascular stent. CheXtra
places masks on both devices. CutLER's proposals localize the surgical clips but struggles more for the stent reaching $0.17$ and $0.25$ IoU, and
its remaining masks partition the thorax without isolating the object. The stent is the
harder case: CheXtra recovers it as multiple fragments rather than one mask but still
marks the device, where CutLER does not. 

\section{Assigning Proposals to Classes with \sysname{}}
\label{sec:curation}

\begin{figure}[tb]
  \centering
  \includegraphics[width=\linewidth]{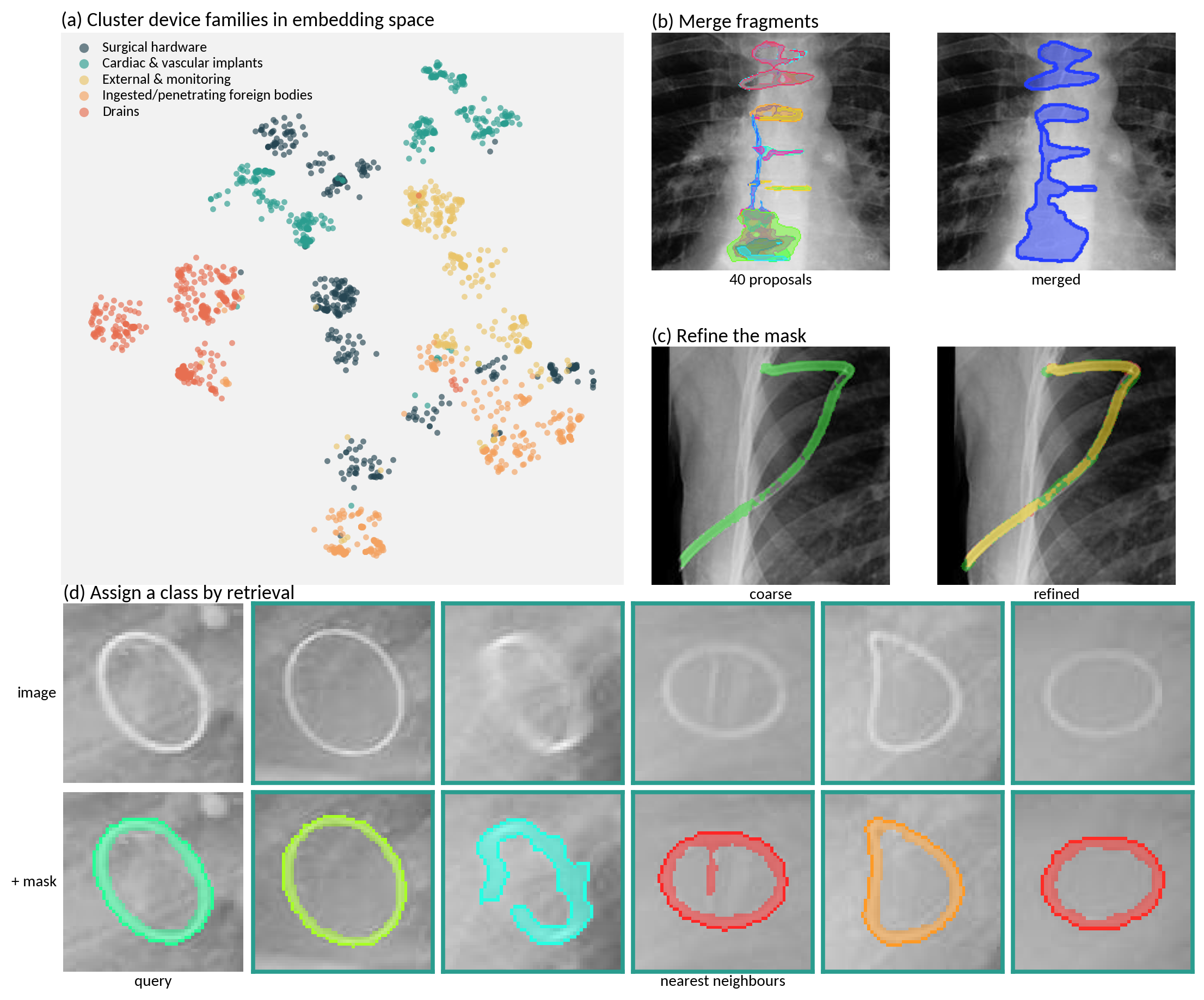}
  \caption{\textbf{The curation loop} on real MIMIC-CXR proposals from the synthetic-trained
  proposer. \textbf{(a)} Instance embeddings cluster proposals, so the curator assigns whole groups
  at once (visualization via h-NNE\cite{sarfraz2022hierarchical}). \textbf{(b)} A \emph{merge} reunites a fragmented device into one instance before
  assignment. \textbf{(c)} A \emph{refine} regrows a thin device along its centreline. \textbf{(d)} A
  \emph{retrieval} pulls in clusters that match a labelled example. Fill colour marks instance
  identity.}
  \label{fig:curation}
\end{figure}

The proposals become annotations only once a person assigns each to a category. \sysname{} clusters the proposals and labels a whole cluster from one
assignment, so the curator can decide over groups rather than single masks (\cref{fig:curation}).

We consider different types of assignments to handle proposed instances. 
A \emph{manual} assignment labels one instance. \emph{Bulk} assignments label a whole cluster at any
level of the FINCH\cite{finch} cluster hierarchy. \emph{Propagate} trains a classifier on the labels already accepted and
proposes the rest using pre-extracted features of each instance. \emph{Retrieve} takes one labelled example and pulls in every cluster that shows
the same content. The curator confirms or overrides each suggestion, and every accepted label records
the mode it came from. A \emph{merge} can then reunite the fragments of a device, which the proposer returns
in pieces (\cref{tab:fragmerge}), and a \emph{refine} allows fixing of a loose boundary.

Clusters separate device families, a merge reassembles a fragmented catheter, a retrieval gathers
matching instances, and the curated masks span cardiac, surgical, ingested, and implanted-port
families (\cref{fig:curation}). We note that nothing in the engine is specific to the domain or to our
proposer. Since it assigns classes over clusters of proposals from any source, the same workflow runs on
other images, in another domain, over different proposal generators like SAM.

\section{Discussion}
\label{sec:discussion}
Most currently existing off-the-shelf proposal mechanisms seem overwhelmed by the general
domain shift and by the structural fine-granularity of foreign objects in chest radiographs.
Across the two unseen real datasets, \methodname{} was the only label-free proposer to provide
reasonable proposals on both. The ablation showed that placement had limited effect beyond
in-body positioning, which can be traced back to the fact that real support devices are
positionally stereotyped, running midline or central, so a canonical layout already matches
them and an anatomical prior adds little once the objects sit on the body. The larger amount of
data had more effect on the model's ability to generate sharper masks than on pure
localization, which could be attributed to the model learning to focus on specific shapes and
becoming better at identifying the expected outlines. The improvement in localization is
strongly influenced by device coverage, which is intuitive, as it supplies the diversity that
the off-the-shelf proposers never saw.

The label-free alternatives fail for one shared reason. Each depends on a signal that the rare
devices do not carry, whether a compact high-contrast blob for salience and unsupervised
objectness, or an in-distribution appearance for generation. A thin, unmentioned,
out-of-distribution device offers none of these, so these routes weaken on the long tail that a
foreign-object dataset most needs to cover. Procedural compositing asks only that the object
can be drawn, which sidesteps these signals. The synthetic-to-real gap remains, and it keeps
such a model from serving as a deployable segmenter~\cite{tobin,tremblay}.

\section{Limitations}
\label{sec:limitations}
Our metrics give only a partial view of the task, because the reference annotations are
incomplete. object-CXR provides boxes and ellipses rather than masks, so we can score it only
at box recall with a loose IoU of 0.3, which bounds how tightly localization can be measured
there. Neither reference set covers the full object space, external objects for one and
catheters and lines for the other, and no complete foreign-object segmentation benchmark
exists.  A faithful score needs independent, exhaustively annotated ground
truth.

The curation engine is a workflow demonstration. We do not measure annotation time or
inter-annotator agreement, which are properties of a multi-annotator deployment rather than of
the method, and belong to a dedicated user study. For the same reason we do not train a
downstream model on the curated masks, as that would test the labels through our own pipeline
rather than against an independent standard. 

\section{Conclusion}
\label{sec:conclusion}
We showed that for foreign-object localization in chest X-rays, where pixel labels are scarce,
synthetic data serves best as a class-agnostic proposal generator that recalls devices across
unseen real domains, and that a human can then turn those proposals into labels through
cluster-level curation. The ablation identified device-appearance diversity as the factor that
carries this transfer, ahead of placement realism and corpus scale. Whether the approach
extends to a complete, independently annotated benchmark, and to domains beyond frontal chest
radiographs, remains open. We release the pipeline and curation engine.

\bibliographystyle{splncs04}
\bibliography{main}

\author{}
\authorrunning{Seibold et al.}
\institute{}

\title{Supplementary Material: Building Foreign Object Segmentation datasets for Chest X-Rays with Synthetic-trained proposals} \titlerunning{Supplementary Material:  Building Foreign Object Segmentation datasets}

\clearpage
\appendix
\renewcommand{\theHsection}{supp.\Alph{section}}

\maketitle
\label{sec:supp}

\section{Generative image models.}
\begin{figure}[p]
  \centering
  \includegraphics[width=0.9\linewidth]{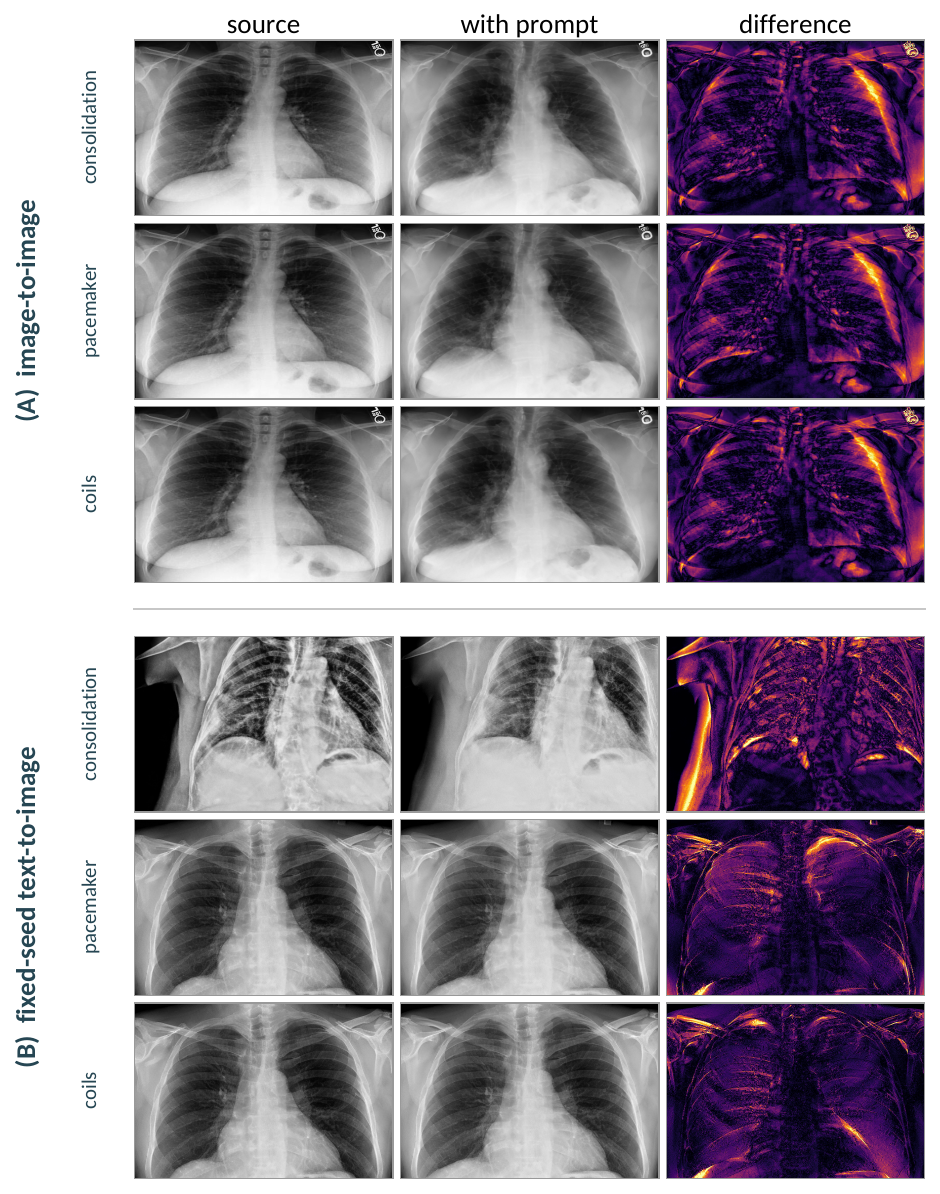}
  \caption{\textbf{A generative chest-X-ray model does not yield a foreign-object mask.} Two routes
  to obtain a device from RoentGen-v2~\cite{roentgenv2}, which is RoentGen~\cite{roentgen} fine-tuned
  on MIMIC-CXR, for a trained finding (consolidation, a positive control) and two devices (pacemaker,
  endobronchial coils). \textbf{(A)} Image-to-image preserves the input and inserts nothing for any
  prompt. \textbf{(B)} Fixed-seed text-to-image renders the trained finding, with consolidation
  appearing over the lung fields, but no device. The source-to-output difference is diffuse rendering
  variation rather than a localized object, and it cannot be thresholded into a mask. The model obeys
  prompts yet places no segmentable foreign object.}
  \label{fig:genfail}
\end{figure}

The main text benchmarks off-the-shelf, unsupervised, and cross-domain proposers. A fourth
label-free route is to generate a device into a radiograph and segment it. \Cref{fig:genfail} shows
why this route fails, with a trained finding as a positive control that the model responds to
prompts. The figure illustrates the mechanism on one generator and does not sweep every generative
configuration. Generation uses RoentGen-v2 at guidance scale $4$, $75$ steps, and $512{\times}512$,
and panel \textbf{(A)} runs image-to-image at denoising strength $0.7$. The complete prompts are:
\begin{itemize}
\item \emph{neutral}: ``Chest radiograph.''
\item \emph{consolidation}: ``Chest radiograph with dense airspace consolidation in the right lower
lobe.''
\item \emph{pacemaker}: ``Chest radiograph with a permanent pacemaker in the left chest and a lead
to the right ventricle.''
\item \emph{endobronchial coils}: ``Chest radiograph with multiple endobronchial lung volume
reduction coils in the upper lobes.''
\end{itemize}

\FloatBarrier
\section{RANZCR box recall (flooding regime).}
\Cref{tab:ranzcrbox} reports RANZCR box recall. A loose box around a catheter is trivially hit, so
volume-heavy proposers, and even Random-box, surpass the synthetic-trained proposer, whose strength
is precise mask localization. This is why the main paper scores RANZCR by mask recall.

\begin{table}[ht]
\caption{RANZCR \emph{box} recall at three IoU thresholds, a flooding-prone regime. Top rows are
in-domain upper bounds; \textbf{bold} is best excluding them. Subscripts are 95\% bootstrap CI
half-widths over images.}
\label{tab:ranzcrbox}
\centering
\scriptsize
\setlength{\tabcolsep}{5pt}
\renewcommand{\arraystretch}{1.25}
\begin{tabular}{@{}l ccc@{}}
\toprule
& \multicolumn{3}{c}{RANZCR (box, $\mathrm{AR}@100$)}\\
\cmidrule(lr){2-4}
trained on $\rightarrow$ eval & IoU 0.3 & IoU 0.5 & IoU 0.7\\
\midrule
object-CXR-real~\cite{objectcxr} \emph{(in-dom.)}    & \cipm{0.846}{.019} & \cipm{0.210}{.024} & \cipm{0.014}{.007}\\
RANZCR-real~\cite{ranzcr} \emph{(in-dom.)}           & \cipm{0.996}{.003} & \cipm{0.962}{.01} & \cipm{0.867}{.018}\\
\cmidrule(lr){1-4}
COCO-Mask2Former~\cite{mask2former}                  & \cipm{0.604}{.027} & \cipm{0.176}{.022} & \cipm{0.018}{.008}\\
COCO-Mask R-CNN~\cite{maskrcnn}                      & \cipm{0.802}{.021} & \cipmb{0.405}{.028} & \cipmb{0.104}{.018}\\
SAM~\cite{sam}                                       & \cipm{0.116}{.018} & \cipm{0.020}{.007} & \cipm{0.008}{.004}\\
SAM-HQ~\cite{samhq}                                  & \cipm{0.288}{.026} & \cipm{0.061}{.013} & \cipm{0.010}{.005}\\
SAM2~\cite{sam2}                                     & \cipm{0.003}{.004} & \cipm{0.000}{.000} & \cipm{0.000}{.000}\\
MedSAM~\cite{medsam}                                 & \cipm{0.848}{.017} & \cipm{0.280}{.024} & \cipm{0.022}{.008}\\
CutLER~\cite{cutler}                                 & \cipmb{0.857}{.017} & \cipm{0.394}{.028} & \cipm{0.040}{.012}\\
\cmidrule(lr){1-4}
Felzenszwalb~\cite{felzenszwalb}                     & \cipm{0.798}{.022} & \cipm{0.188}{.023} & \cipm{0.015}{.007}\\
Selective Search~\cite{selectivesearch}              & \cipm{0.755}{.023} & \cipm{0.309}{.027} & \cipm{0.041}{.011}\\
Felz$\rightarrow$MedSAM~\cite{felzenszwalb,medsam}   & \cipm{0.770}{.023} & \cipm{0.172}{.022} & \cipm{0.014}{.007}\\
Random-box                                           & \cipm{0.764}{.024} & \cipm{0.173}{.021} & \cipm{0.014}{.007}\\
\cmidrule(lr){1-4}
\methodname{} (ours)                                       & \cipm{0.478}{.029} & \cipm{0.161}{.02} & \cipm{0.035}{.01}\\
\bottomrule
\end{tabular}
\end{table}

\FloatBarrier

\begin{figure}[t]
  \centering
  \includegraphics[width=\linewidth]{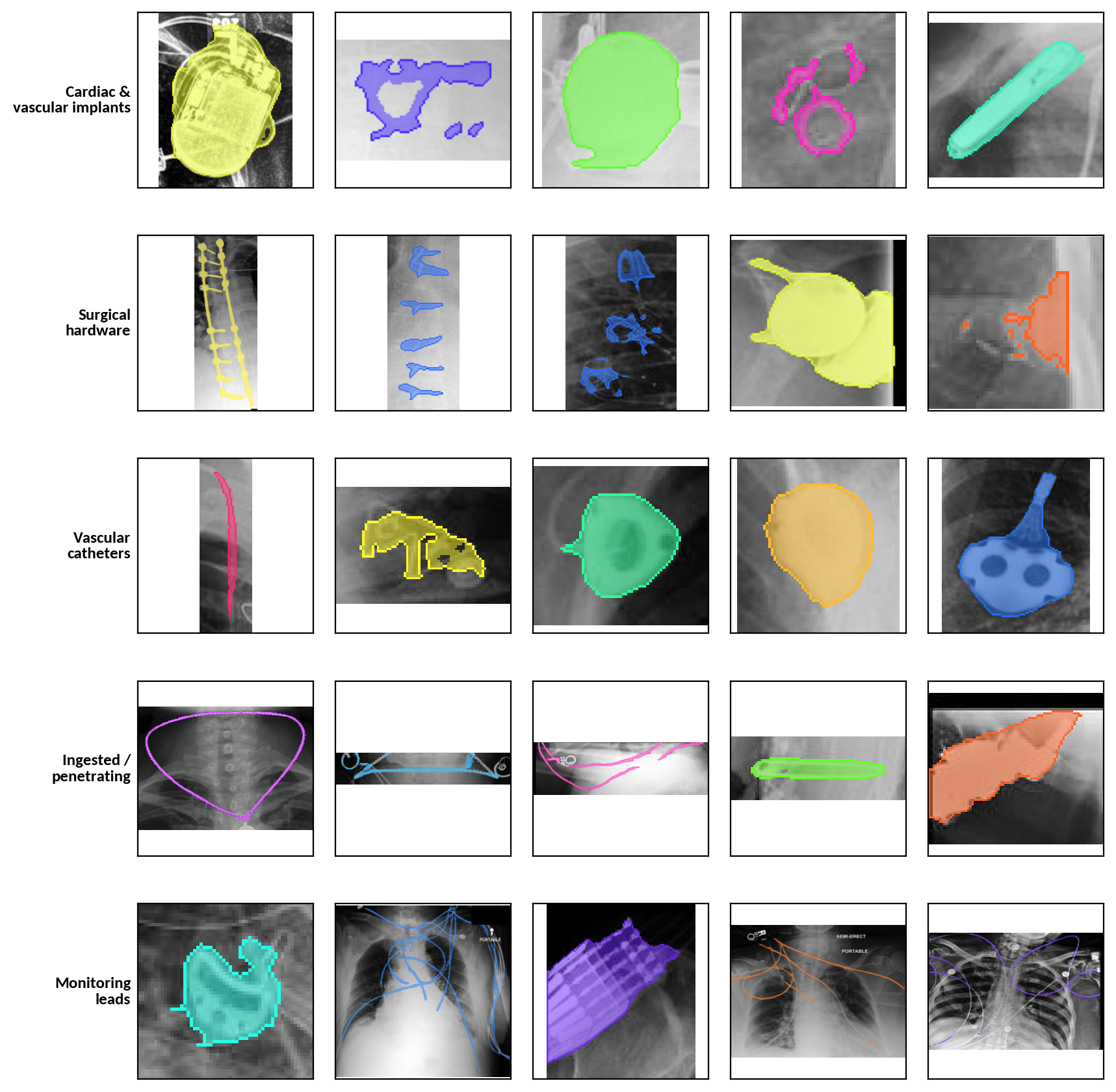}
  \caption{Curated real instances, grouped by device family, one row per family and one fill colour
  per instance.}
  \label{fig:gallery}
\end{figure}

\section{Synthetic generation rules.}
The generator follows fixed rules. Every object is a
procedural draw or a cut-paste, so its mask is exact by construction and no step depends on a
learned generative model.

\noindent\textbf{Per-image sampling.}
Each background draws a scene layout. With probability \num{0.85} it takes one of \num{23} clinical
archetypes (\cref{tab:archetypes}), and otherwise it samples freely. An archetype emits its
\emph{anchoring} devices, adds each optional co-device with even odds, then tops up with a set
number of \emph{extra} objects. ICU-complex, for instance, adds one to three, and a free image
draws one to twelve. Both the extra draws and the free-image draws sample the \num{119}-type
library uniformly, not by clinical prevalence, so a rare implant appears as often as a common line.
About a fifth of images take the \emph{no-finding} archetype and carry no object, which supplies
device-free hard negatives. A further sixth take \emph{minimal-labels} and carry only burned-in
text or laterality markers, which teach the proposer to ignore acquisition furniture.

\begin{table}[t]
\caption{The \num{23} clinical archetypes and their sampling weights. An image draws an
archetype with probability \num{0.85}. The archetype emits its \emph{anchoring} devices, adds
each optional co-device with even odds (omitted here for space), then tops up with
$n_\text{extra}$ objects drawn uniformly from the \num{119}-type library. Weights are relative
and normalised at sampling. \emph{No-finding} and \emph{minimal-labels} together leave about two
in five images device-free or marker-only, as hard negatives.}
\label{tab:archetypes}
\centering
\scriptsize
\setlength{\tabcolsep}{4pt}
\renewcommand{\arraystretch}{1.1}
\begin{tabular}{@{}l l c c@{}}
\toprule
Archetype & Anchoring device(s) & $n_\text{extra}$ & wt.\\
\midrule
no-finding        & ---                                         & 0    & 0.25\\
minimal-labels    & burned-in text / markers                    & 0--1 & 0.18\\
icu-complex       & ET tube, tunneled CVC, 5-lead monitor       & 1--3 & 0.06\\
chronic-pacemaker & pacemaker                                   & 0--1 & 0.05\\
post-CABG         & sternotomy wires                            & 0--2 & 0.05\\
port-chemo        & chest port                                  & 0--2 & 0.05\\
arrest-intubated  & ET tube, 5-lead monitor, cables             & 0--2 & 0.04\\
dialysis          & dialysis catheter                           & 0--2 & 0.04\\
ICD               & ICD generator + lead                        & 0--1 & 0.03\\
oncology-brachy   & brachytherapy seeds                         & 0--2 & 0.03\\
spine-surgery     & spinal rods                                 & 0--2 & 0.03\\
pediatric-ingest  & (optional ingested items)                   & 1--2 & 0.03\\
CRT               & CRT generator + leads                       & 0--1 & 0.02\\
TAVR              & valve mesh stent                            & 0--1 & 0.02\\
ECMO              & ECMO cannulae, ET tube                      & 0--3 & 0.02\\
trauma            & pellet cluster                              & 0--3 & 0.02\\
breast-recon      & breast implant                              & 0--2 & 0.02\\
neuro-shunt       & VP shunt                                    & 0--1 & 0.02\\
pleural-effusion  & chest tube                                  & 0--2 & 0.02\\
LVAD              & LVAD body + cannulae                        & 0--2 & 0.01\\
cancer-lung       & ablation antenna                            & 0--1 & 0.01\\
IABP              & intra-aortic balloon pump                   & 0--2 & 0.01\\
pulmonary-emboli  & embolization coils                          & 0--1 & 0.01\\
\bottomrule
\end{tabular}
\end{table}

\noindent\textbf{Per-category multiplicity.}
Each category then draws an instance count that respects clinical multiplicity. Implanted bodies
and airway devices appear once, one pacemaker and one ET tube. Markers and repeated hardware draw a
small count: surgical clips one to five, burned-in text one to four, vertebroplasty cement one to
three, chest tubes and drains one to two. The generator places, scales, rotates, and renders each
instance independently.

\noindent\textbf{Laterality priors.}
A few devices carry a left/right prior, under the radiograph convention that the patient's left is
image-right. Pacemakers, ICDs, and CRT bodies anchor to the upper-left chest \num{95}\% of the
time. Chest ports anchor to the patient-right \num{80}\% of the time, and tunneled or dialysis
lines \num{70}\%. Other categories carry no lateral bias.

\noindent\textbf{Placement.}
The generator anchors positions on anatomy by default. A chest-anatomy segmenter (CXAS) parses the
background into clavicle, rib, spine, and mediastinal masks, and each rule samples its anchor inside
the relevant structure: a pacemaker below the clavicle, spinal rods along the spine, a chest tube at
the pleural margin. An archetype's devices then co-locate as they do in practice. Two ablation modes
decouple this prior without changing appearance. \emph{Fallback} drops the anatomy masks, so each
rule reverts to a fixed geometric default. \emph{Random} shifts the whole anatomy stack by a random
per-device offset, which keeps multi-part devices coherent while decorrelating their position from
the patient. 

\noindent\textbf{X-ray-aware rendering.}
The material class of each instance sets its rendering mode. Solid implant bodies use a
translucent-implant or screen blend, a bright rim with anatomy showing through the interior.
Radio-opaque tubes and thin wires use Beer--Lambert attenuation or bone-occlusion darkening, so a
lead dims where it crosses a rib. Burned-in text uses a crisp opaque overlay, and skin-surface
objects use a soft-edge feather. The generator randomises contrast and opacity per instance and
gamma-sharpens partial-alpha edges, which removes the soft halo of a paste. It writes only inserted
pixels, so the masked and unmasked images differ on the object alone and yield a pixel-exact mask at
no labeling cost.

\noindent\textbf{Base pool.}
Backgrounds are device-free frontal and lateral MIMIC-CXR films. We select them by pairing
report-text filtering with a RAD-DINO~\cite{raddino} image-side classifier that flags residual,
report-unmentioned hardware. On lateral views the generator withholds
frontal-only categories.

\begingroup
\makeatletter
\@fileswfalse
\let\hyper@anchorstart\@gobble
\let\hyper@anchorend\relax
\bibliography{main}
\endgroup

\end{document}